\documentclass[10pt,twocolumn,letterpaper]{article}

\usepackage{dicta}
\usepackage{times}
\usepackage{epsfig}
\usepackage{algorithm2e}
\usepackage[export]{adjustbox}
\usepackage{array}
\usepackage{textcomp}
\usepackage{stfloats}
\usepackage{url}
\usepackage{verbatim}
\usepackage{graphicx}
\usepackage{cite}
\usepackage{xcolor}
\usepackage{booktabs}
\usepackage{amssymb}
\usepackage{amsmath,amsfonts,bm}
\usepackage{multirow}
\usepackage{csquotes}
\usepackage{paralist}
\usepackage{threeparttable}

\usepackage[normalem]{ulem}

\usepackage{algpseudocode}

\dictafinalcopy
\def\dictaPaperID{}

\usepackage[pagebackref=true,breaklinks=true,letterpaper=true,colorlinks,bookmarks=false]{hyperref}

\begin{document}

\title{IGME: Efficient Chained Method Ensemble for Transferable Semantic Segmentation Attacks}

\author{
Mengqi He \qquad Jing Zhang\\[0.5ex]
\small School of Computing, College of Systems \& Society\\
\small The Australian National University, Canberra, ACT, Australia\\[0.5ex]
\small \texttt{Mengqi.He@anu.edu.au} \quad
\texttt{Jing.Zhang@anu.edu.au}
}
\date{}

\hypersetup{
  pdftitle={IGME: Efficient Chained Method Ensemble for Transferable Semantic Segmentation Attacks},
  pdfauthor={Mengqi He and Jing Zhang}
}

\maketitle
\thispagestyle{plain}

\begin{abstract}
Semantic segmentation models are vulnerable to transferable adversarial perturbations, yet evaluating transfer attacks on dense prediction models can be computationally expensive. Existing ensemble attacks often rely on multiple surrogate models, increasing the computation cost, even harder for segmentation. This paper studies an efficient single-source alternative for transferable attacks on semantic segmentation. We formulate transferable attack composition as a chained computation over differentiable attack components, allowing the expensive source-model gradient computation to be shared. To reduce the update instability introduced by chained composition, we further use an integrated-gradient-style path-averaged direction as an empirical stabilization heuristic. Experiments on Pascal VOC and Cityscapes evaluate the resulting transferability efficiency trade-off across CNN- and transformer-based segmentation models. IGME achieves competitive transferability compared with single-source baselines and favorable runtime compared with model-ensemble attacks, while requiring access to only one source model.
\end{abstract}

\section{Introduction}

Semantic segmentation is a core dense-prediction task used in applications such as autonomous driving and medical image analysis~\cite{chen2022vision,li2021semantic}. Despite strong clean-image accuracy, segmentation models remain vulnerable to small adversarial perturbations~\cite{xie2017adversarial,arnab2018robustness}. The dense and spatially structured output makes this vulnerability particularly important: a perturbation can alter large regions of a prediction while remaining visually subtle.

\begin{figure}[t]
    \centering
  \begin{tabular}{{c@{ } c@{ } c@{ }}}
{\includegraphics[width=0.32\linewidth]{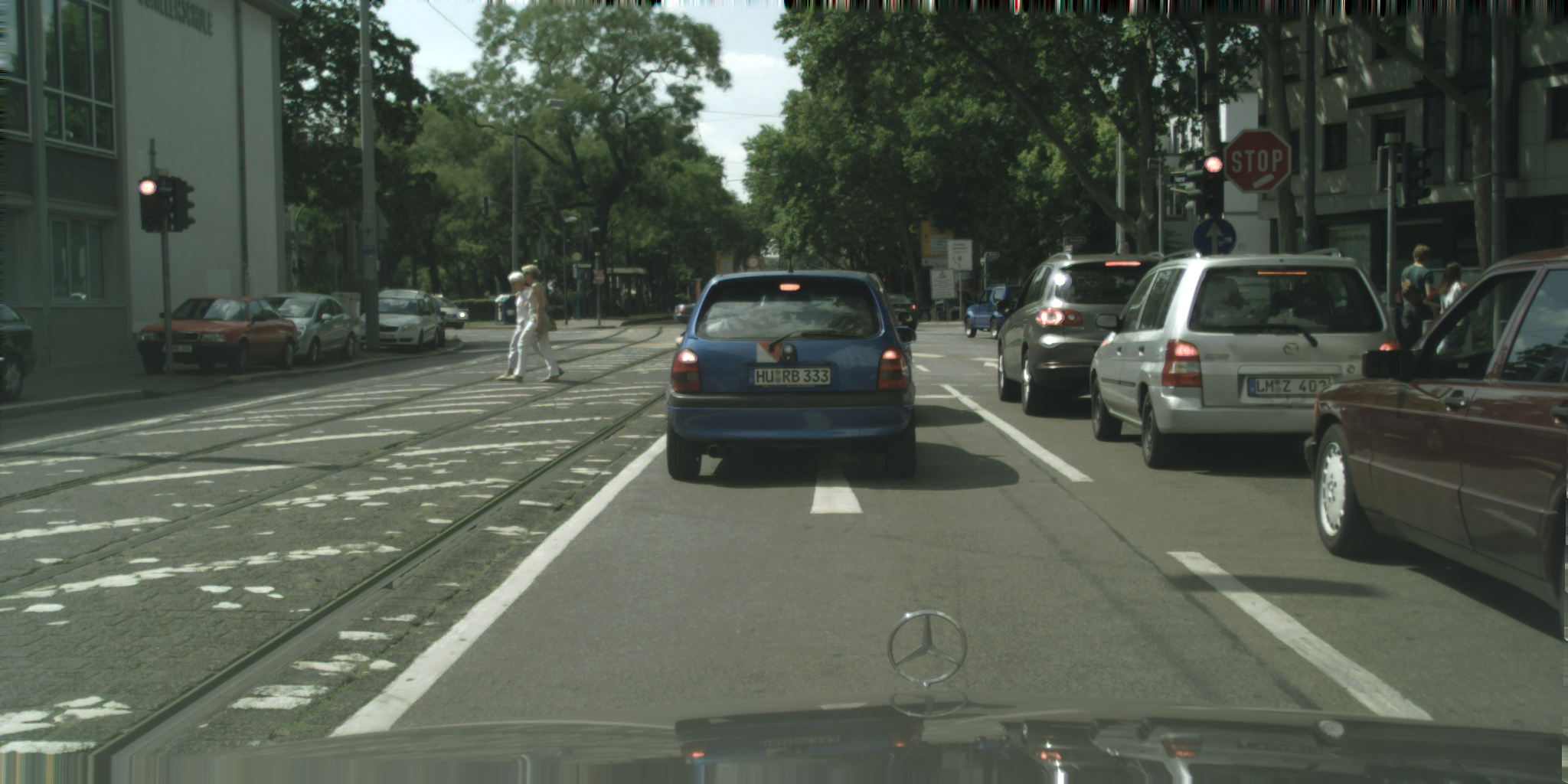}}&  
{\includegraphics[width=0.32\linewidth]{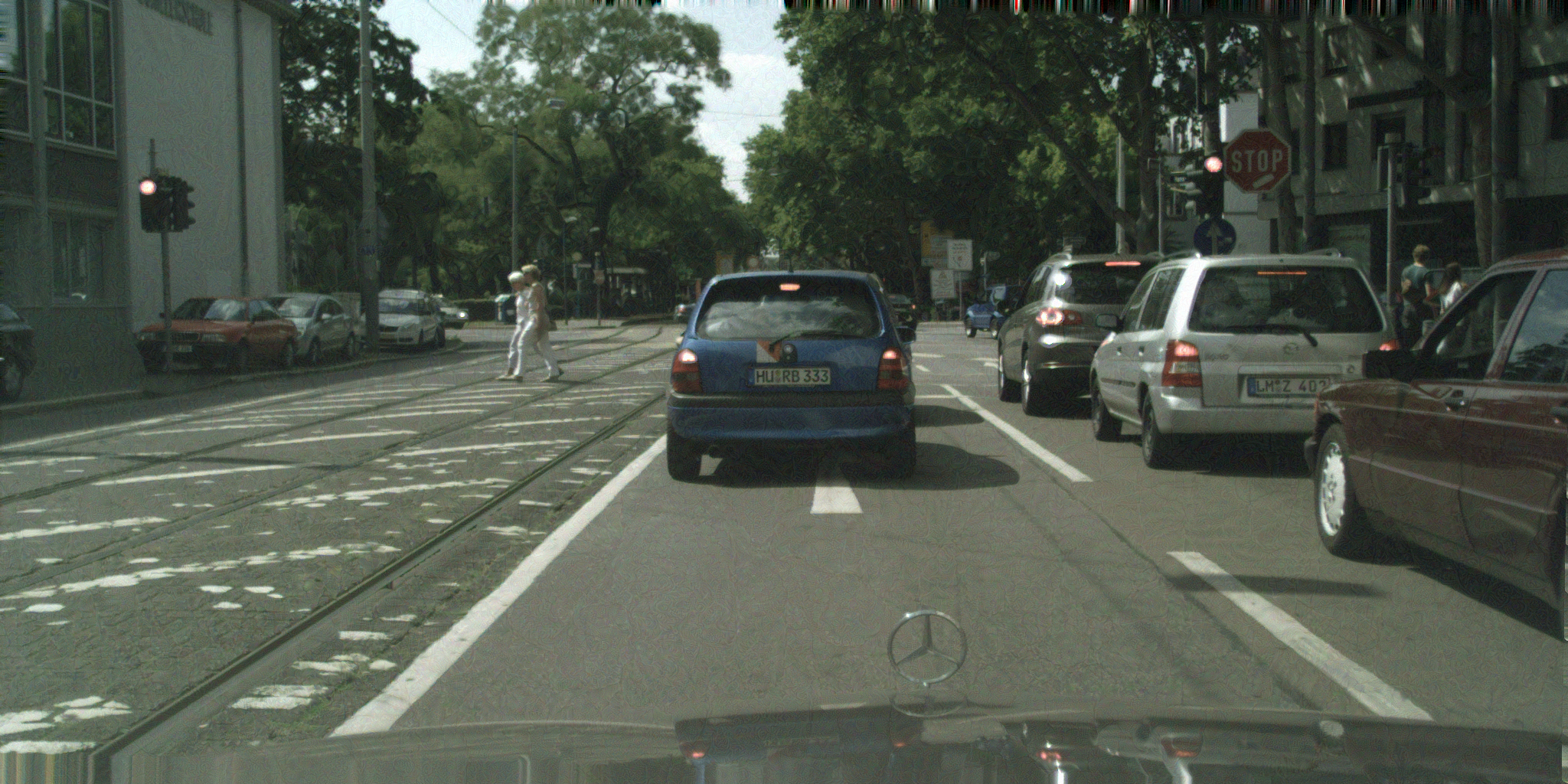}}&  
{\includegraphics[width=0.32\linewidth]{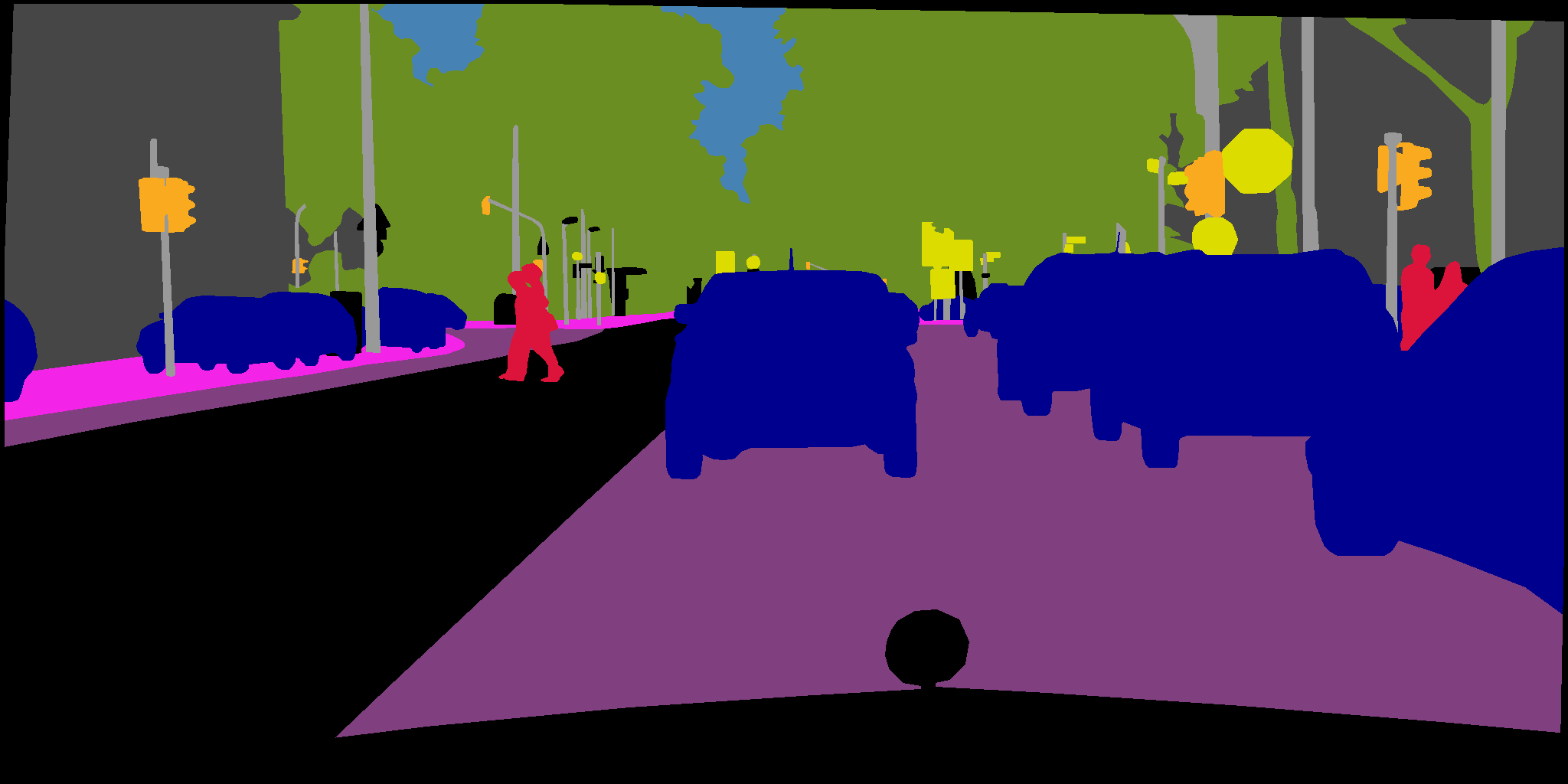}}\\ 
\footnotesize{Image}&\footnotesize{Adv}&\footnotesize{GT}\\
{\includegraphics[width=0.32\linewidth]{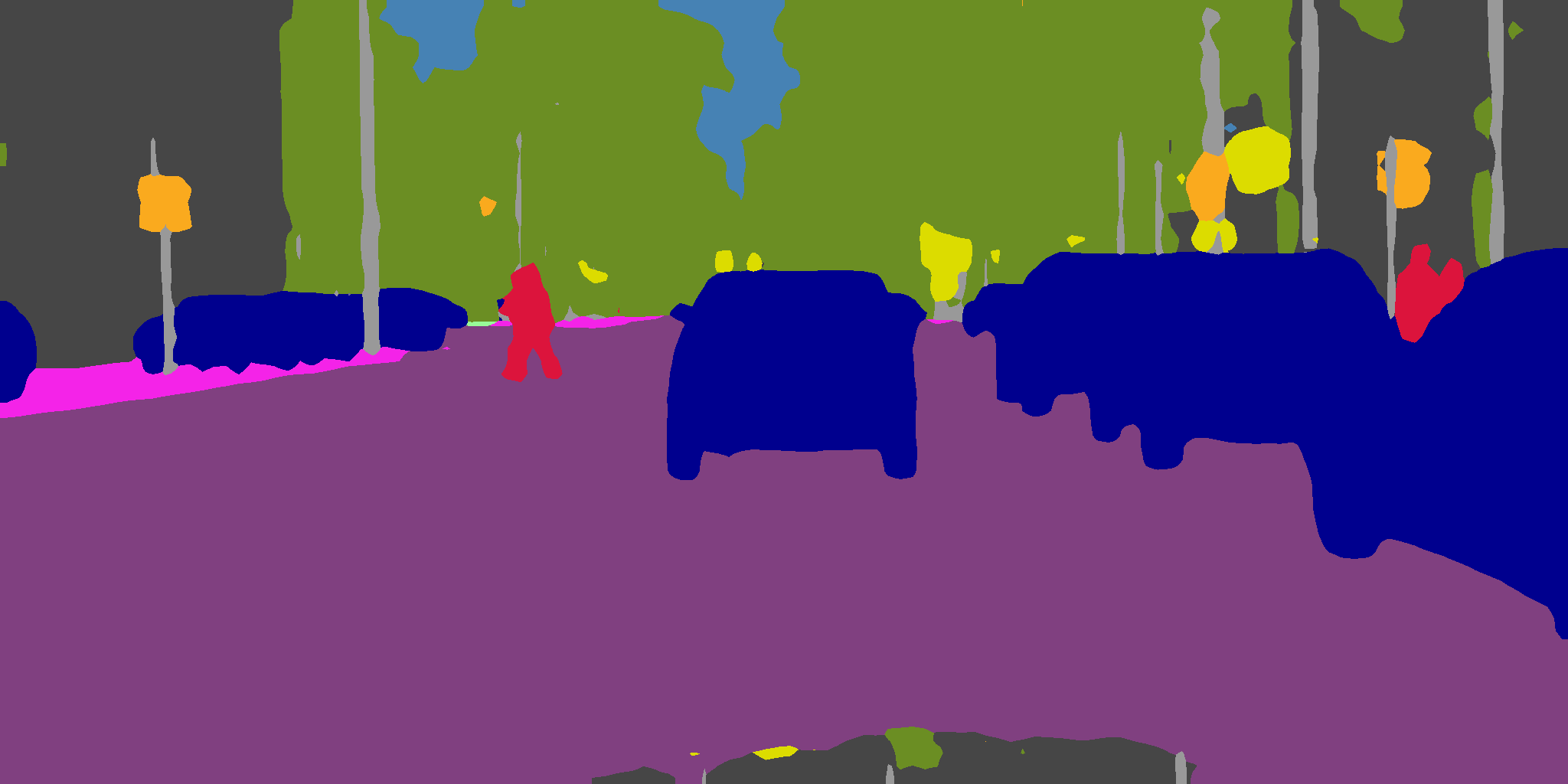}}&  
{\includegraphics[width=0.32\linewidth]{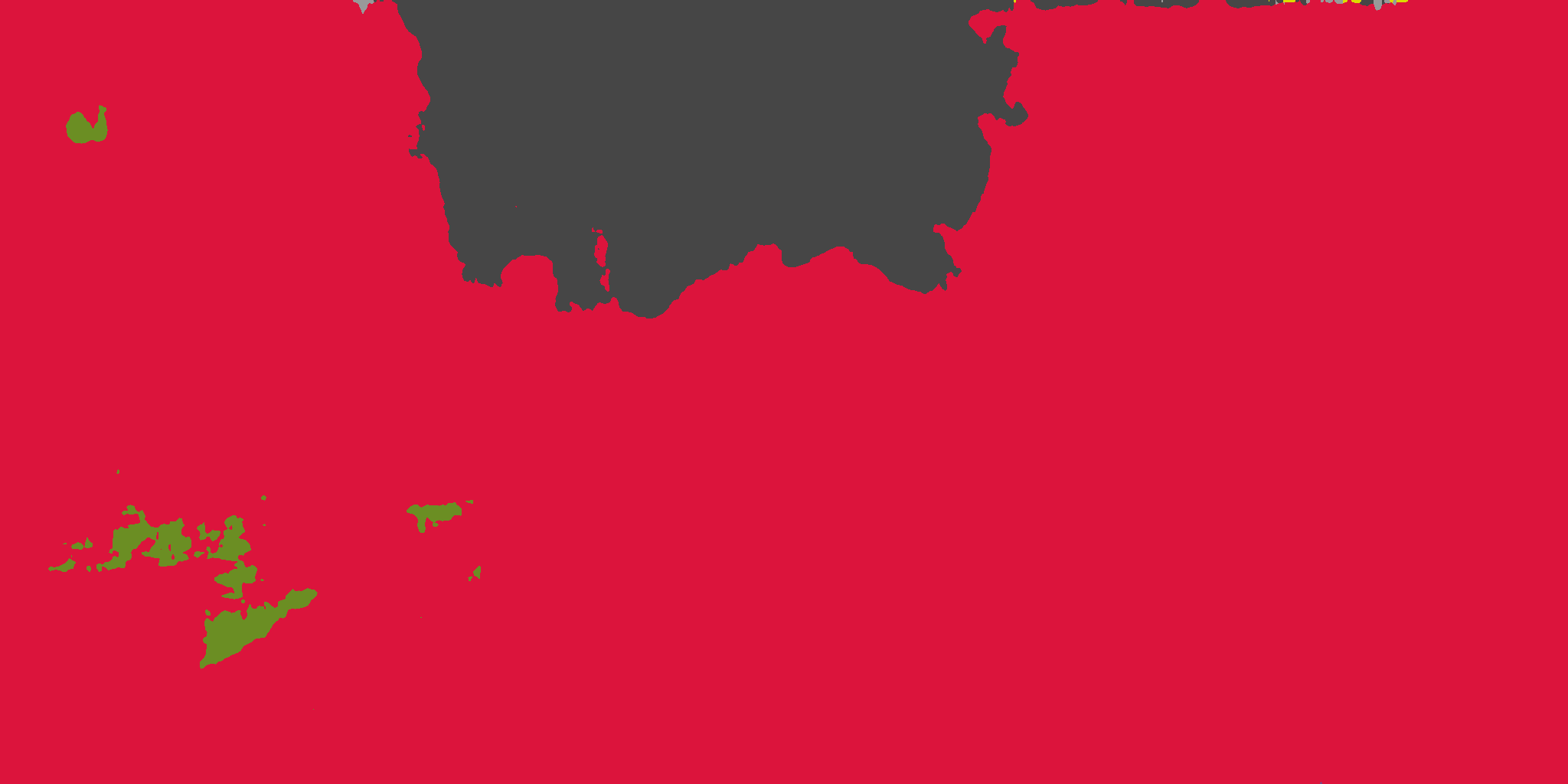}}&  
{\includegraphics[width=0.32\linewidth]{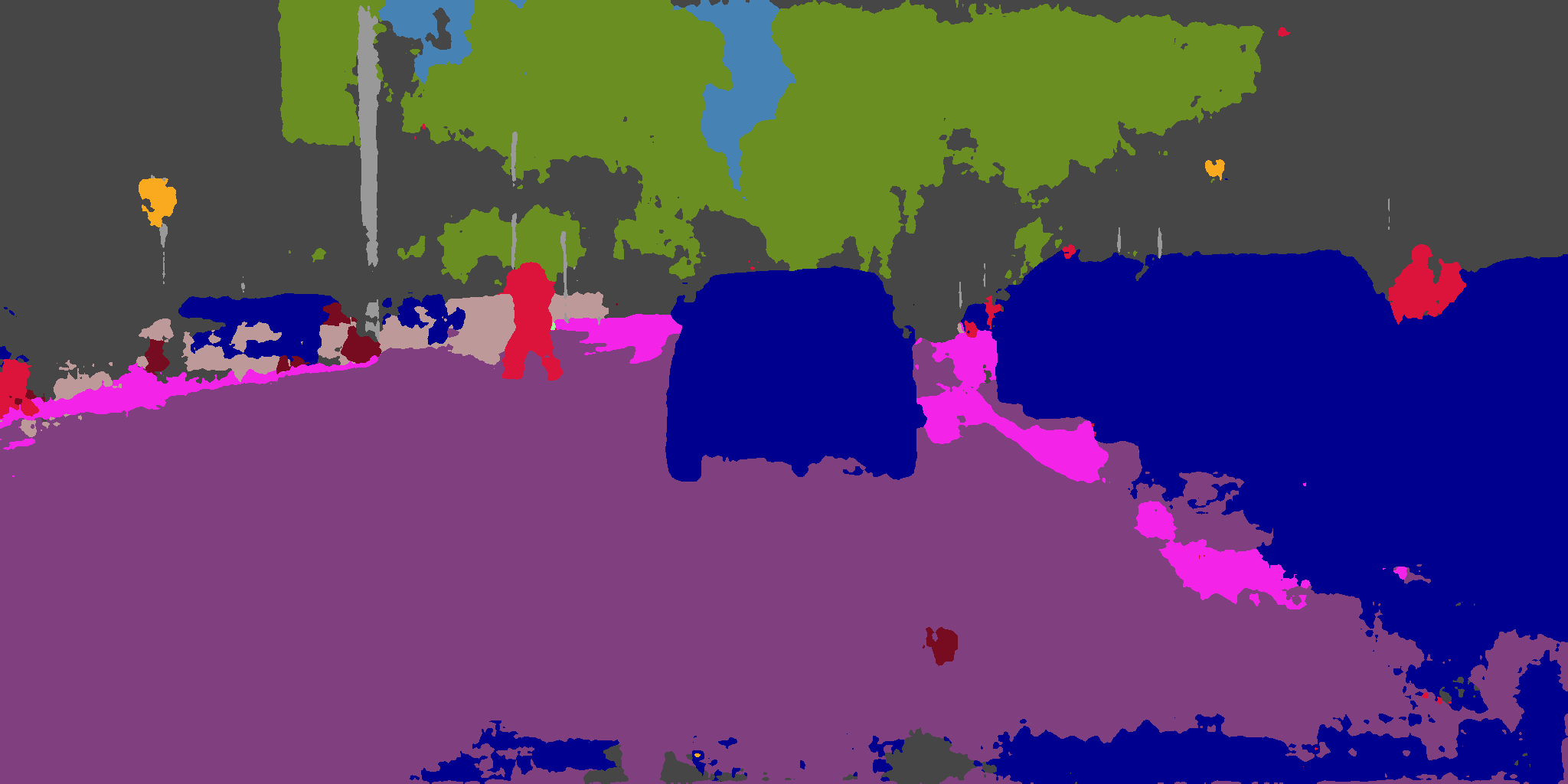}}\\
\footnotesize{Clean}&\footnotesize{Noise}&\footnotesize{Noise+}\\
   \end{tabular}
    \caption{\textbf{Motivating source-to-target transfer example.} A PGD adversarial example is generated on DeepLabV3 using a Cityscapes image and evaluated on DeepLabV3+~\cite{chen2017rethinking,chen2018encoder,Cordts2016Cityscapes,PGD_attack}. The source prediction (\enquote{Noise}) is more strongly degraded than the target prediction (\enquote{Noise+}), illustrating the architecture-transfer gap. The example follows the evaluation protocol of prior transferable-segmentation work~\cite{he2024transferable} and is included only to motivate the efficiency problem studied here.}
    \label{semantic_seg_towards_attack}
\end{figure}

Transferable attacks consider a more constrained setting than ordinary white-box evaluation. An adversarial example is generated on an accessible source model and then applied to target models whose parameters or gradients are unavailable. Figure~\ref{semantic_seg_towards_attack} illustrates this source-to-target gap: the example substantially degrades the source-model prediction, but its effect is weaker after transfer to a model with a different architecture. 
Prior work on transferable attacks for semantic segmentation evaluated classification-oriented techniques under dense-prediction objectives~\cite{he2024transferable}.  Among the strategies used to improve transferability, ensemble attacks are the most direct: aggregating information from multiple surrogate models, or from multiple attack branches applied to one model, generally transfers better than any single attack alone~\cite{liu2016delving,dong2018boosting,DBLP:conf/cvpr/XiongLZH022}. 

However, this strength comes at a cost that is especially severe for segmentation. Model ensembles require a separate forward backward pass through the source model for every surrogate. Because segmentation losses and gradients are computed densely over every pixel rather than a single class score, each such pass is already far more expensive than in classification, so this repeated computation is especially costly for dense prediction. Method ensembles avoid the need for multiple models, but a naive implementation still evaluates each attack branch independently, so the source-model computation is still repeated once per branch. 

We therefore study chained attack-component composition in a single-source model setting. Instead of generating and aggregating several independent attack branches, differentiable transfer-enhancing components are placed in one computation graph so that the source-model computation is shared. Because chained composition may produce inconsistent update directions, we further introduce an integrated-gradient-style path average as a stabilization heuristic. We refer to the resulting method as IGME.

Our contributions are summarized as follows:
\begin{itemize}
    \item We analyze the cost of model ensembles and independently evaluated attack-component ensembles, and formulate a chained component composition that reduces repeated source-model gradient evaluations.
    \item We introduce IGME, which uses an integrated-gradient-style path-averaged direction to stabilize chained attack-component updates.
    \item We evaluate the on Pascal VOC and Cityscapes across CNN- and transformer-based segmentation models, together with runtime, composition, and preliminary SAM analyses.
\end{itemize}

\section{Related Work}

\noindent\textbf{Adversarial attacks on semantic segmentation.}
Adversarial attacks search for bounded input perturbations that degrade a model's prediction~\cite{Explaining_and_Harnessing_Adversarial_Examples,PGD_attack}. Dense prediction introduces additional structure because the objective is aggregated over many spatial locations. Early work evaluates the robustness of segmentation models with standard gradient attacks~\cite{arnab2018robustness}, while DAG directly manipulates dense outputs~\cite{xie2017adversarial} and SegPGD reweights correctly and incorrectly classified pixels during iterative optimization~\cite{gu2022segpgd}. Other studies consider universal or targeted perturbations~\cite{hendrik2017universal,adversarial_example_semantic_seg_iclr2017workshop}, proximal optimization~\cite{DBLP:conf/cvpr/RonyPA23}, uncertainty-aware losses~\cite{DBLP:conf/wacv/MaagF24}, and certified-radius-guided objectives~\cite{DBLP:conf/eurosp/QuLW23}. 
These methods establish that CNN- and transformer-based segmentation models are vulnerable, but most are designed primarily for white-box attack.

\noindent\textbf{Transferable adversarial attacks.}
Transfer attacks aim to preserve adversarial effect when an example crafted on a source model is evaluated on an unseen target. Representative approaches improve transfer through input diversity~\cite{CihangXie2018ImprovingTO}, translation-invariant gradient processing~\cite{YinpengDong2019EvadingDT}, momentum or Nesterov-style optimization~\cite{dong2018boosting,JiadongLin2019NesterovAG}, variance tuning~\cite{XiaosenWang2021EnhancingTT}, and surrogate or feature refinement~\cite{zhou2018transferable,AdityaGaneshan2019FDAFD,naseer2021improving}. For semantic segmentation, Dynamic Scaling links transferability to the multi-scale behavior of segmentation architectures~\cite{gu2021adversarial}. A prior systematic study adapts NI, DI, TI, IAA, DAG, and SegPGD to a shared dense-prediction protocol~\cite{he2024transferable}. We retain that test protocol for comparable baselines.

Integrated Gradients accumulate gradients along a path from a baseline to an input~\cite{DBLP:conf/icml/SundararajanTY17}. TAIG uses integrated-gradient directions for transferable classification attacks~\cite{DBLP:conf/iclr/HuangK22}, and MIG combines path integration with momentum to attack both convolutional and transformer classifiers~\cite{ma2023transferable}. Our use of path integration has a narrower role: it is an empirical stabilizer for the direction produced by chained attack components. 

\noindent\textbf{Model ensembles and method ensembles.}
Aggregating several surrogate models can improve transferability by combining predictions, logits, losses, or gradients~\cite{liu2016delving,dong2018boosting}. SVRE reduces gradient variance across an ensemble~\cite{DBLP:conf/cvpr/XiongLZH022}, while adaptive weighting and sharpness-aware objectives further improve model-ensemble attacks~\cite{DBLP:conf/iccv/ChenYCCL23,chen2024rethinking}. Self-ensemble attacks avoid storing several independently trained models, but can still require multiple augmented branch evaluations~\cite{DBLP:conf/cvpr/HuangCCWZ23}. We instead chains differentiable attack components in one source-model graph.


\section{Method}
\label{sec:ensemble_attack}
Ensemble attacks often improve transferability by aggregating information from multiple surrogate models, but repeated model evaluations are expensive for dense prediction~\cite{liu2016delving,DBLP:conf/cvpr/XiongLZH022,chen2024rethinking}. Self-ensemble attacks reduce the number of distinct surrogates, yet may still evaluate multiple augmented branches~\cite{DBLP:conf/cvpr/HuangCCWZ23}. We instead consider a single-source method ensemble in which differentiable transfer-enhancing components share one computation graph.

We use \enquote{method ensemble} in the adversarial-attack sense, but the chained objects in our formulation are differentiable attack components rather than complete iterative optimizers. Examples include differentiable input transformations, smoothing operators, and gradient-shaping modules. Complete attacks such as PGD, NI, or SegPGD may include sign operations, clipping, projection, and stateful iteration; they are not treated as ordinary differentiable functions unless explicitly unrolled.

The chained formulation is a heuristic composition of differentiable attack operators. It is not a probabilistic ensemble and is not an algebraic approximation to averaging perturbations, losses, logits, or gradients. Its purpose is to share the expensive source-model computation while allowing several transfer-enhancing components to influence the update direction. IGME contains two stages: chained component composition, followed by an IG-style path average that empirically stabilizes the resulting direction.

\subsection{Threat Model and Transferability}
\label{sec:threat_model}
Let $g_s$ be an accessible source segmentation model, let $g_t$ denote an unseen target model, and let $(\mathbf{x},\mathbf{y})$ be a clean image label pair. Given perturbation budget $\epsilon$, an attack algorithm $A$ produces $\mathbf{x}_{\mathrm{adv}}=A(g_s,\mathbf{x},\mathbf{y};\epsilon)$ with $\|\mathbf{x}_{\mathrm{adv}}-\mathbf{x}\|_\infty\leq\epsilon$. The attack does not use target-model parameters or gradients. We measure its transfer effect on $g_t$ through a segmentation discrepancy $D$, $\mathrm{Trans}(A;g_s,g_t,\mathbf{x})=D\!\left(g_t(\mathbf{x}_{\mathrm{adv}}),\mathbf{y}\right)$. Using the scalar segmentation loss as $D$ and averaging over a set of target models $\mathcal{G}_t$ and the data distribution $\mathcal{X}$ gives the average transfer objective
\begin{equation}
\label{eq_transferability_ensemble}
\mathrm{Trans}_{\mathrm{avg}}(A;g_s,\mathcal{G}_t)=
\mathbb{E}_{g_t\sim\mathcal{G}_t,(\mathbf{x},\mathbf{y})\sim\mathcal{X}}
\left[\mathcal{L}_{\mathrm{seg}}\!\left(g_t(\mathbf{x}_{\mathrm{adv}}),\mathbf{y}\right)\right].
\end{equation}
In the experiments, lower adversarial mIoU and higher normalized mIoU drop (SR) indicate a stronger transferred attack.

\subsection{Complexity Analysis of Ensemble Attacks}

Let $A$ denote the cost of applying one differentiable attack component, transformation, or gradient-shaping operator, and let $B$ denote one source-model forward--backward pass. Suppose an independently evaluated component ensemble contains $N$ branches. A naive average-gradient implementation requires approximately $O(NA+NB)$ per attack iteration because every branch is evaluated through the source model. A model ensemble similarly requires repeated model evaluations, with simplified cost $O(A+NB)$.

For chained component composition without path averaging, the components share one source-model forward--backward pass, giving a per-iteration cost of $O(NA+B)$. Over $K$ attack iterations, the cost is $O(K(NA+B))$, and for IGME with $M$ path samples per iteration, the cost becomes $O(KM(NA+B))$.
The $M$ path samples may be processed as a batch to improve hardware utilization, but the number of model-equivalent gradient evaluations still scales with $M$. In the default setting, $K=10$ and $M=2$, so IGME performs 20 source-model gradient evaluations and uses twice the path-gradient evaluations of the chained baseline. This is consistent with Table~\ref{tab:runtime}, where IGME is approximately $2.20\times$ and $2.64\times$ slower than chained composition on DeepLabV3 and FCN, respectively. The ratios need not equal two exactly because batching, memory allocation, path construction, and fixed tensor operations introduce additional overhead.

\begin{figure}[t!]
   \begin{center}
   \begin{tabular}{{c@{ }}}

{\includegraphics[width=0.95\linewidth]{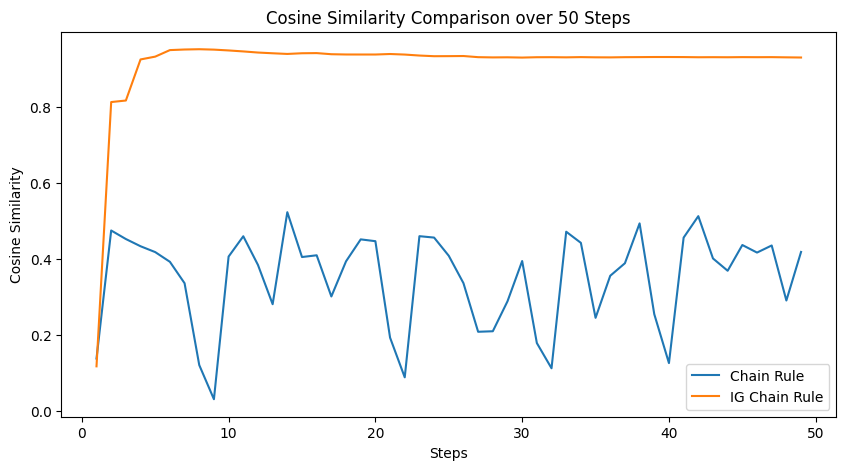}}\\
   \end{tabular}
   \end{center}
    \caption{\textbf{Cosine similarity for chained component attacks with and without IG-style path averaging.} Higher similarity indicates more stable update directions in this experiment; it is used as empirical evidence of step-wise stabilization rather than as a formal guarantee of transferability.} 
    \label{fig:cossim_vs}
\end{figure}

\noindent\textbf{Efficient Transferable Attack with Integrated Gradient Method Ensemble.}\quad
Despite the runtime advantage of chained component composition, its update direction can be less stable when applied directly. To improve the empirical stability of the update, we use path-averaged gradient information, motivated by prior work showing that stabilized gradients can improve transferability~\cite{JiadongLin2019NesterovAG}.
The integrated gradient method~\cite{DBLP:conf/icml/SundararajanTY17} computes an integral of gradients along the path from a baseline to the input. We use this idea as a path-averaging heuristic for chained attack components~\cite{DBLP:conf/iclr/HuangK22}.
Semantic segmentation produces dense logits, so the IG direction must be defined with respect to a scalar attack objective rather than directly with respect to the dense model output. Let \(\mathcal{T}=T_1\circ T_2\circ\cdots\circ T_N\) denote the differentiable component chain and define
\begin{equation}
\label{eq_scalar_seg_objective}
    J_t(\mathbf{z}) =
    \mathcal{L}_{\mathrm{seg}}\left(g(\mathcal{T}(\mathbf{z})),\mathbf{y}\right),
\end{equation}
where \(\mathcal{L}_{\mathrm{seg}}\) is the scalar segmentation loss. Formally, the IG-style path-averaged direction is:
\begin{equation}
\label{integrate_gradient}
    \mathbf{d}_t =
    (\mathbf{x}^{adv}_t-\mathbf{r})\odot
    \int_0^1
    \nabla_{\mathbf{z}}J_t\!\left(\mathbf{r}+\eta(\mathbf{x}^{adv}_t-\mathbf{r})\right)
    \,d\eta .
\end{equation}
where $\mathbf{r}$ is the baseline image, chosen as a black image in this setting, and \(\odot\) denotes element-wise multiplication. In implementation, the integral is approximated using \(M\) path samples:
\begin{equation}
\label{eq_igme_discrete}
    \widehat{\mathbf{d}}_t =
    (\mathbf{x}^{adv}_t-\mathbf{r})\odot
    \frac{1}{M}
    \sum_{m=1}^{M}
    \nabla_{\mathbf{z}}J_t\!\left(\mathbf{r}+\frac{m}{M}(\mathbf{x}^{adv}_t-\mathbf{r})\right).
\end{equation}
With the default $M=2$, Eq.~\ref{eq_igme_discrete} averages gradients at the midpoint and endpoint of the straight-line path from the baseline to the current adversarial image.
Under this definition, the integration averages gradient information along the path. We use this path averaging as a stabilization heuristic, rather than as a proof of optimality or guaranteed transferability. Empirically, the cosine similarity between consecutive IG-style update directions is higher than that of the regular gradient-based method in our reported experiment, as shown in Fig.~\ref{fig:cossim_vs}.

We therefore use the path average only as an empirical stabilizer of the chained direction. Its effect on transferability and runtime is evaluated separately in Section~\ref{sec_experiments}.
The update is:
\begin{equation}
\label{eq_ni_di_ti_adversarial_attack}
    \begin{aligned}
        \mathbf{x}^{adv}_{t+1} =
        \mathrm{Clip}_{\mathbf{x},\epsilon}
        \left(
        \mathbf{x}^{adv}_t+\alpha\,\mathrm{sign}(\widehat{\mathbf{d}}_t)
        \right).
    \end{aligned}
\end{equation}
where \(\mathrm{Clip}_{\mathbf{x},\epsilon}\) denotes projection to the \(L_\infty\) ball around the clean image \(\mathbf{x}\), followed by clipping to the valid image range, and $\alpha$ denotes the step size.
We use $M=2$ in all reported IGME experiments.
In the end, the algorithm can be formulated as:
\SetKwComment{Comment}{/* }{ */}
\SetKwInput{KwInput}{Input}                
\SetKwInput{KwOutput}{Output}              
\SetAlgoLined

\begin{algorithm}
\caption{Integrated Gradients Method Ensemble (IGME)}\label{alg:igme}
\KwInput{Clean image $\mathbf{x}$, label $\mathbf{y}$, source model $g$, differentiable components $T_1,\ldots,T_N$, segmentation loss $\mathcal{L}_{\mathrm{seg}}$, baseline $\mathbf{r}$, perturbation budget $\epsilon$, step size $\alpha$, iterations $K$, path samples $M$}
\KwOutput{Adversarial example $\mathbf{x}_K^{adv}$}
\SetAlgoLined
$\mathbf{x}_0^{adv}\gets \mathbf{x}$\;
$\mathcal{T}\gets T_1\circ T_2\circ\cdots\circ T_N$\;
\For{$t=0$ \KwTo $K-1$}{
    $J_t(\mathbf{z})\gets \mathcal{L}_{\mathrm{seg}}(g(\mathcal{T}(\mathbf{z})),\mathbf{y})$\;
    $\widehat{\mathbf{d}}_t\gets 0$\;
    \For{$m=1$ \KwTo $M$}{
        $\mathbf{z}_{t,m}\gets \mathbf{r}+\frac{m}{M}(\mathbf{x}_t^{adv}-\mathbf{r})$\;
        $\widehat{\mathbf{d}}_t\gets \widehat{\mathbf{d}}_t+\nabla_{\mathbf{z}}J_t(\mathbf{z}_{t,m})$\;
    }
    $\widehat{\mathbf{d}}_t\gets (\mathbf{x}_t^{adv}-\mathbf{r})\odot \widehat{\mathbf{d}}_t/M$\;
    $\mathbf{x}_{t+1}^{adv}\gets \Pi_{\mathcal{B}_{\infty}(\mathbf{x},\epsilon)\cap[0,1]}(\mathbf{x}_t^{adv}+\alpha\,\mathrm{sign}(\widehat{\mathbf{d}}_t))$\;
}
\end{algorithm}

IGME is designed to improve the transferability efficiency trade-off rather than to dominate every target architecture. The experiments therefore compare it with strong single-source attacks and with more expensive model-ensemble baselines, while separately measuring the benefit of path averaging over chained composition.

\section{Experiments}
\label{sec_experiments}
In this section, we systematically evaluate the proposed adversarial attacks for semantic segmentation, focusing on their transferability and computational efficiency.
To comprehensively evaluate these aspects of our proposed ensemble attack, we conduct a series of experiments, comparing our method with several strong baselines, analyzing parameter sensitivity, and assessing computational complexity.

\subsection{Settings}
We follow the transferable semantic-segmentation evaluation protocol of prior work~\cite{he2024transferable}.

\noindent\textbf{Dataset.}
 We train our models on the augmented Pascal VOC 2012~\cite{pascal-voc-2012} dataset, which contains 21 classes with 10,582 images. During training, we apply online data augmentation, including random scaling within the range of $[0.5, 2]$, random cropping to 513x513 pixels, and random horizontal flipping. For generalization, we also train on the Cityscapes~\cite{Cordts2016Cityscapes} dataset, containing 3,475 images with 19 classes. We use random cropping to 768x768 pixels, color jitter, and random horizontal flipping for data augmentation. Evaluation is performed on their respective test sets: 1,449 images for Pascal VOC 2012 (resized to 513x513) and 1,525 images for Cityscapes (original resolution).

\noindent\textbf{Evaluation Metrics.}
We evaluate segmentation performance using mean IoU ($\text{mIoU}$). Peak Signal-to-Noise Ratio (PSNR) and Structural Similarity Index Measure (SSIM)~\cite{wang2004image} quantify image fidelity rather than transferability itself. Success rate ($\text{Sr}$) is defined as $S_r = 1 - \frac{\text{mIoU}_\text{adv}}{\text{mIoU}_\text{clean}}$, where $\text{mIoU}_\text{adv}$ and $\text{mIoU}_\text{clean}$ are the segmentation performances of adversarial and clean images, respectively.

\noindent\textbf{Adversarial Attacks.}
For single model methods, classic adversarial attacks like FGSM~\cite{Explaining_and_Harnessing_Adversarial_Examples}, a single-step gradient-based attack, and PGD~\cite{PGD_attack}, a multi-step iterative attack, are evaluated. Specialized semantic segmentation attacks such as SegPGD~\cite{gu2022segpgd}, adapted from PGD for segmentation tasks, and DAG~\cite{xie2017adversarial}, targeting the segmentation output space, are also considered. For transferability, advanced methods including NI~\cite{JiadongLin2019NesterovAG}, leveraging Nesterov accelerated gradients, DI~\cite{CihangXie2018ImprovingTO}, enhancing transferability through diverse input transformations, and TI~\cite{YinpengDong2019EvadingDT}, using translation-invariant perturbations, are evaluated.

For model ensemble methods, the Ensemble (ENS) attack~\cite{liu2016delving}, combining multiple models to enhance attack performance, and the Stochastic Variance Reduced Ensemble (SVRE) attack~\cite{DBLP:conf/cvpr/XiongLZH022}, employing variance reduction strategies for improved attack effectiveness, are utilized.

\paragraph{Parameter setting.}
We use an $L_\infty$ perturbation budget of $\epsilon=8/255$ and $K=10$ attack iterations for all iterative attacks unless otherwise stated. IGME uses $M=2$ path samples per attack iteration in Algorithm~\ref{alg:igme}. Therefore, the default IGME setting performs $KM=20$ source-model gradient evaluations. The two path samples correspond to $\eta\in\{1/2,1\}$ in Eq.~\ref{eq_igme_discrete}, and are processed as a batch when memory permits. Runtime results are measured under the same image resolution, batch size, and perturbation budget for the compared methods.

\begin{table*}[t!]
  \centering
  \scriptsize
  \renewcommand{\arraystretch}{0.95}
  \renewcommand{\tabcolsep}{1.8mm}
  \caption{\textbf{Quantitative results on Pascal VOC.} Lower adversarial mIoU and higher SR indicate stronger attacks. Bold denotes the best values within each target column.}
  \begin{tabular}{l|c|cc|cc|cc|cc|cc}
  \toprule[1.1pt]
  &&\multicolumn{2}{c|}{Image Quality}&\multicolumn{8}{c}{Target (performance is evaluated with $\text{mIoU}\downarrow$ (left) and $\text{SR}\uparrow$ (right))}\\ \hline
  Source & Attack & $\text{PSNR}\uparrow$ & $\text{SSIM}\uparrow$ & \multicolumn{2}{c|}{DV3Res50} & \multicolumn{2}{c|}{DV3Res101}& \multicolumn{2}{c|}{FCNVGG16}& \multicolumn{2}{c}{Mask2Former}  \\ \hline
  \multirow{9}{*}{DV3Res50}  & FGSM~\cite{Explaining_and_Harnessing_Adversarial_Examples} & 30.18 & 0.5098 & 0.3400 & 0.5480 & 0.5134 & 0.3410 &  0.4628 & 0.2890 & 0.6617 & 0.1238 \\ 
  & PGD~\cite{PGD_attack} & 39.70 & 0.7814 & 0.0938 & 0.8753 & 0.4232 & 0.4568 &  0.5681 & 0.1273  & 0.6930 & 0.0823  \\ 
   \cline{2-12}
  & SegPGD~\cite{gu2022segpgd} & 40.13 & 0.7897 & 0.1127 & 0.8502 & 0.5659 & 0.2736 &   0.6030 & 0.0740& 0.6998 & 0.0734  \\ 
   & DAG~\cite{xie2017adversarial} & \textbf{46.09} & 0.9340 & 0.1268 & 0.8314 & 0.6271 & 0.1951 & 0.5977 & 0.0819 & 0.7288 & 0.0349 \\ 
   \cline{2-12}
      &  NI~\cite{JiadongLin2019NesterovAG} & 32.84 & 0.5912 &0.0861 & 0.8855 & 0.2610  &0.6650 &  0.4246 & 0.3478  & 0.5926 & 0.2153 \\ 
  & DI~\cite{CihangXie2018ImprovingTO} & 39.70 & 0.7793 & 0.0950 & 0.8737 &  0.2929 & 0.6241 &  0.5035 & 0.2266  & 0.6241 & 0.1736 \\ 
  & TI~\cite{YinpengDong2019EvadingDT} & 39.26 & 0.7808 & 0.0997 & 0.8675 & 0.3871 & 0.5031 & 0.5193 & 0.2023 & 0.6395 & 0.1532 \\
    & IAA ~\cite{zhu2022rethinking} & 35.69 & \textbf{0.9396} & 0.0823 & 0.8897 & 0.5379 & 0.3095 & 0.5625 & 0.1360  & 0.6386 & 0.1543 \\
    & IGME(ours) & 32.91 & 0.6025 & \textbf{0.0758} & \textbf{0.8992} & \textbf{0.1892} & \textbf{0.7571} & \textbf{0.2100} & \textbf{0.6774} & \textbf{0.3859} & \textbf{0.4889} \\
   \midrule[0.7pt]
   \multirow{9}{*}{DV3Res101} & FGSM~\cite{Explaining_and_Harnessing_Adversarial_Examples} & 30.17  & 0.5104 & 0.4959 & 0.3407 & 0.3774 & 0.5156 &  0.4764 & 0.2682 & 0.7047 & 0.0669 \\ 
   & PGD~\cite{PGD_attack} &  39.64  & 0.7804 & 0.3564 & 0.5262 & 0.1001  & 0.8715 & 0.5673 & 0.1286 & 0.7085 & 0.0618  \\ 
   \cline{2-12}
  & SegPGD~\cite{gu2022segpgd} & 40.09 & 0.7891 & 0.4557 & 0.3941 & 0.1131 & 0.8548 & 0.5985 & 0.0806 & 0.7153 & 0.0528  \\
   & DAG~\cite{xie2017adversarial} & \textbf{46.41} & \textbf{0.9401} & 0.5710 & 0.2409 & 0.1569 & 0.7986 & 0.6057 & 0.0696 & 0.7294 & 0.0342 \\ 
   \cline{2-12}
    &  NI~\cite{JiadongLin2019NesterovAG} & 32.83 & 0.5925 & 0.2244 &0.7017 &0.0922 & 0.8817 & 0.4341 & 0.3332 & 0.6032 & 0.2013 \\ 
    & DI~\cite{CihangXie2018ImprovingTO} & 39.70 & 0.7791 & 0.2545 & 0.6617 & 0.0973 & 0.8751 &  0.5029 & 0.2275 & 0.6428 & 0.1488  \\ 
    & TI~\cite{YinpengDong2019EvadingDT} & 39.26 & 0.7805 & 0.3543 & 0.5290 & 0.1068  & 0.8629 & 0.5235& 0.1959 & 0.6538 & 0.1343 \\
  
    & IAA ~\cite{zhu2022rethinking} & 34.21& 0.9244 & 0.4597 & 0.3888 & 0.1315 & 0.8312  & 0.5690 & 0.1259 & 0.6437 & 0.1476 \\
& IGME(ours) & 33.74 & 0.7834 & \textbf{0.1976} & \textbf{0.7373} & \textbf{0.0755} & \textbf{0.9031} & \textbf{0.3744} & \textbf{0.4249} & \textbf{0.4035} & \textbf{0.4657} \\
\midrule[0.7pt]
  \multirow{8}{*}{FCNVGG16} & FGSM~\cite{Explaining_and_Harnessing_Adversarial_Examples} & 30.18 & 0.5099 & 0.3811 & 0.4933 & 0.4641 & 0.4043 &  0.1715 & 0.7366 & 0.7235 & 0.0420  \\ 
  & PGD~\cite{PGD_attack} & 36.73 & 0.5106  & 0.1989 & 0.7356 &  0.2981  & 0.6174  & 0.0309 & 0.9525 & 0.7209  & 0.0454  \\ 
   \cline{2-12}
  & SegPGD~\cite{gu2022segpgd} & 37.86 & 0.6023 & 0.3346 & 0.5552 &  0.4337 & 0.4433 & 0.0608 & 0.9066& 0.7328 & 0.0297  \\
   & DAG~\cite{xie2017adversarial}  & \textbf{42.64} & 0.9216 & 0.4723 & 0.3721 & 0.5538 & 0.2892 & 0.1378 & 0.7883 & 0.7258 & 0.0389 \\
    \cline{2-12}
   &  NI~\cite{JiadongLin2019NesterovAG} & 32.75 & 0.6018 & 0.1607 & 0.7864 & 0.2233 & 0.7134 &  0.0326 & 0.9499& 0.6348 & 0.1594   \\ 
    & DI~\cite{CihangXie2018ImprovingTO} & 37.39 & 0.7492 & 0.1747 & 0.7677 & 0.2536  & 0.6745 &  0.0335 & 0.9485 & 0.6631 & 0.1220  \\ 
    & TI~\cite{YinpengDong2019EvadingDT} & 36.35 & 0.7382 & 0.2272 & 0.6980 & 0.3262  & 0.5813 & 0.0320 & 0.9508& 0.6820 & 0.0969 \\
    & IAA ~\cite{zhu2022rethinking} & 34.77& \textbf{0.9566} & 0.4465 & 0.4064 & 0.4325 & 0.4448  & 0.0332 & 0.9490 & 0.6625 & 0.1227  \\
    & IGME(ours) & 33.56 & 0.7455 & \textbf{0.1346} & \textbf{0.8211} & \textbf{0.1942} & \textbf{0.7508} & \textbf{0.0296} & \textbf{0.9545}  & \textbf{0.3928} & \textbf{0.4799} \\
\midrule[0.7pt]
\multirow{8}{*}{Mask2Former} & FGSM~\cite{Explaining_and_Harnessing_Adversarial_Examples} & 30.38 & 0.5199 & 0.3911 & 0.4801 & 0.4741 & 0.3913 & 0.3883 & 0.4036 & 0.4522 & 0.4012  \\ 
  & PGD~\cite{PGD_attack} & 36.83 & 0.5206  & 0.2089 & 0.7223 &  0.3081  & 0.6045  & 0.4623 & 0.2900 & 0.1223 & 0.8381  \\ 
   \cline{2-12}
  & SegPGD~\cite{gu2022segpgd} & 37.96 & 0.6123 & 0.3446 & 0.5418 &  0.4437 & 0.4305 & 0.4878 & 0.2510 & 0.1419 & 0.8121  \\
   & DAG~\cite{xie2017adversarial}  & \textbf{42.74} & \textbf{0.9316} & 0.4823 & 0.3588 & 0.5638 & 0.2762 & 0.4936 & 0.2418 & 0.1369 & 0.8187\\
    \cline{2-12}
   &  NI~\cite{JiadongLin2019NesterovAG} & 32.85 & 0.6118 & 0.1707 & 0.7731 & 0.2333 & 0.7006 &  0.3538 & 0.4566 & 0.0852 & 0.8872  \\ 
    & DI~\cite{CihangXie2018ImprovingTO} & 37.49 & 0.7592 & 0.1847 & 0.7545 & 0.2636  & 0.6617 &  0.4099 & 0.3703 & 0.0875 & 0.8841  \\ 
    & TI~\cite{YinpengDong2019EvadingDT} & 36.45 & 0.7482 & 0.2372 & 0.6846 & 0.3362  & 0.5685 & 0.4266 & 0.3448 & 0.1143 & 0.8486 \\
    & IGME(ours) & 32.86 & 0.5980 & \textbf{0.0720} & \textbf{0.9043} & \textbf{0.1986} & \textbf{0.7451} & \textbf{0.3051} & \textbf{0.5314} & \textbf{0.0638} & \textbf{0.9155} \\
\bottomrule[1.1pt]
  \end{tabular}
  \label{tab:pascalvoc_adversarial_attack}
\end{table*}

\begin{table*}[t!]
  \centering
  \footnotesize
  \renewcommand{\arraystretch}{1.0}
  \renewcommand{\tabcolsep}{2.0mm}
  \caption{\textbf{Ensemble comparison on Pascal VOC 2012~\cite{pascal-voc-2012}.} Lower adversarial mIoU and higher SR indicate stronger attacks. Bold and underlined values denote the best and second-best values within each target column, respectively.}
  \begin{tabular}{l|c|cc|cc|cc|cc}
  \toprule[1.1pt]
  &&\multicolumn{2}{c|}{Image Quality}&\multicolumn{6}{c}{Target (performance is evaluated with $\text{mIoU}\downarrow$ and $\text{SR}\uparrow$)}\\ \hline
  Source & Attack & $\text{PSNR}\uparrow$ & $\text{SSIM}\uparrow$ & \multicolumn{2}{c|}{DV3Res50} & \multicolumn{2}{c|}{DV3Res101}& \multicolumn{2}{c}{FCNVGG16}  \\ \hline
  \multirow{4}{*}{DV3Res50} 
      &  NI~\cite{JiadongLin2019NesterovAG} & 32.84 & 0.5912 & 0.0861 & 0.8855 & 0.2610  &0.6650 & 0.4246 & 0.3478  \\ 
    & ENS~\cite{liu2016delving} & 32.89 & 0.6050 & 0.0799 & 0.8938 & 0.2581 & 0.6687 & 0.3003 & 0.5387 \\
   & SVRE~\cite{DBLP:conf/cvpr/XiongLZH022} & 32.92 & 0.6082 & \textbf{0.0722} & \textbf{0.9040} & \underline{0.2490} & \underline{0.6804} & \underline{0.2856} & \underline{0.5613} \\
  & IGME(ours) & 32.91 & 0.6025 & \underline{0.0758} & \underline{0.8992} & \textbf{0.1892} & \textbf{0.7571} & \textbf{0.2100} & \textbf{0.6774}\\
  \midrule
  \multirow{4}{*}{Mask2Former} 
      &  NI~\cite{JiadongLin2019NesterovAG} & 32.87 & 0.5958 & 0.1707 & 0.7731 & 0.2333 & 0.7006 & 0.3538 & 0.4566 \\ 
    & ENS~\cite{liu2016delving} & 32.91 & 0.6070 & 0.0785 & 0.8956 & 0.2562 & 0.6712 & \underline{0.2987} & \underline{0.5411} \\
   & SVRE~\cite{DBLP:conf/cvpr/XiongLZH022} & 32.94 & 0.6102 & \textbf{0.0718} & \textbf{0.9045} & \underline{0.2472} & \underline{0.6827} & \textbf{0.2824} & \textbf{0.5662} \\
  & IGME(ours) & 32.86 & 0.5980 & \underline{0.0720} & \underline{0.9043} & \textbf{0.1986} & \textbf{0.7451} & 0.3051 & 0.5314 \\
  \bottomrule[1.1pt]
  \end{tabular}
  \label{tab:pascalvoc_adversarial_attack_esm}
\end{table*}

\noindent\textbf{Network structures.}
For Pascal VOC, we evaluate DeepLabV3 with a ResNet-50 backbone (\enquote{DV3Res50}), DeepLabV3+ with ResNet-101 (\enquote{DV3Res101}), FCN with VGG-16 (\enquote{FCNVGG16}), and Mask2Former~\cite{chen2017rethinking,chen2018encoder,long2015fully,cheng2022masked}. The preliminary Cityscapes ensemble study (supplementary material) uses the same checkpoint names. Table~\ref{tab:backbone_parameter_numbers} reports the parameter counts of the three representative CNN segmentation models; Mask2Former is omitted because its parameter count depends on the selected backbone.

\begin{table}[t!]
    \caption{Parameter counts of the representative CNN segmentation models (millions).}
    \label{tab:backbone_parameter_numbers}
    \centering
    \footnotesize
    \setlength{\tabcolsep}{1.5mm}{
        \begin{threeparttable}
        \begin{tabular}{cccc}
        \toprule[1.1pt]
         & DV3Res50 & DV3Res101 & FCNVGG16 \\
        \midrule
        \#Parameters & 23.5 & 42.5 & 134.5 \\
        \bottomrule[1.1pt]
        \end{tabular}
        \end{threeparttable}
    }
\end{table}

\subsection{Experimental Results}
\noindent\textbf{Transferability.}
The full Pascal VOC matrix in Table~\ref{tab:pascalvoc_adversarial_attack}. 

Table~\ref{tab:pascalvoc_adversarial_attack_esm} provides the more direct ensemble comparison. When DV3Res50 is the source, IGME reduces the target mIoU to 0.1892 on DV3Res101 and 0.2100 on FCNVGG16, outperforming the reported single-source NI values on these targets. ENS and SVRE remain competitive or stronger on some source-target pairs, but they evaluate multiple surrogate models. The central empirical claim is therefore an improved transferability efficiency trade-off rather than uniform dominance over every architecture.

\begin{figure}[t]
    \centering
     \begin{tabular}{{r@{ } c@{ } c@{ }c@{ } c@{ }c@{ }c@{ }}}
{\scriptsize Baseline }\hspace{-0.2em} & 
\adjustbox{valign=c}{\includegraphics[width=0.13\linewidth]{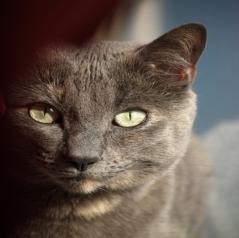}}&  
\adjustbox{valign=c}{\includegraphics[width=0.13\linewidth]{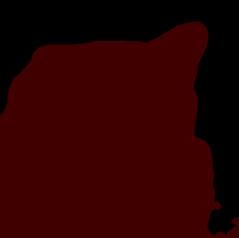}}& 
\adjustbox{valign=c}{\includegraphics[width=0.13\linewidth]{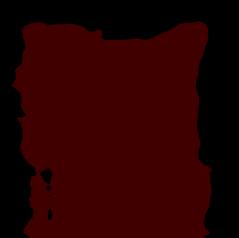}}&  
\adjustbox{valign=c}{\includegraphics[width=0.13\linewidth]{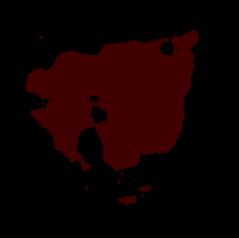}}&  
\adjustbox{valign=c}{\includegraphics[width=0.13\linewidth]{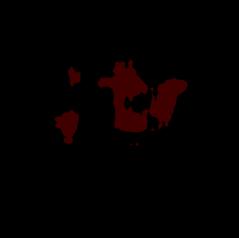}}& 
\adjustbox{valign=c}
{\includegraphics[width=0.13\linewidth]{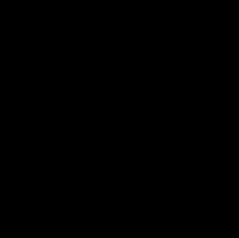}}\\
{\scriptsize \textbf{IGME} }\hspace{-0.2em} &
\adjustbox{valign=c}{\includegraphics[width=0.13\linewidth]{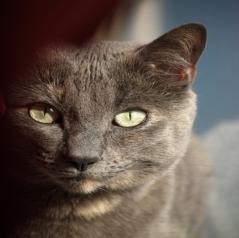}}&  
\adjustbox{valign=c}{\includegraphics[width=0.13\linewidth]{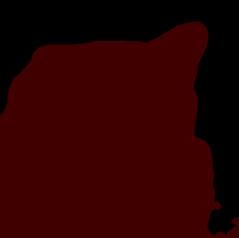}} & 
\adjustbox{valign=c}{\includegraphics[width=0.13\linewidth]{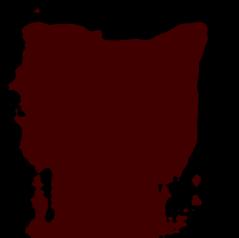}}&  
\adjustbox{valign=c}{\includegraphics[width=0.13\linewidth]{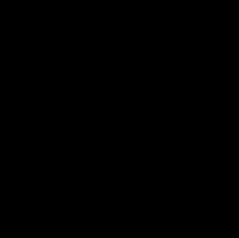}}&  
\adjustbox{valign=c}{\includegraphics[width=0.13\linewidth]{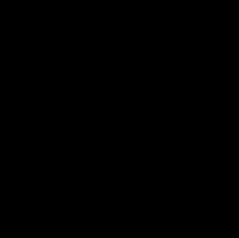}}&  
\adjustbox{valign=c}{\includegraphics[width=0.13\linewidth]{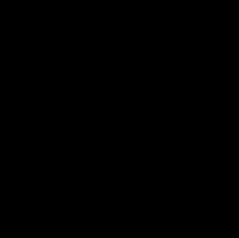}}\\
{ \scriptsize Baseline }\hspace{-0.2em} &
\adjustbox{valign=c}{\includegraphics[width=0.13\linewidth]{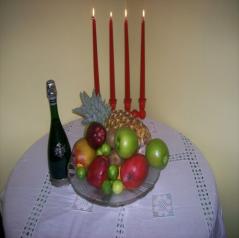}}&  
\adjustbox{valign=c}{\includegraphics[width=0.13\linewidth]{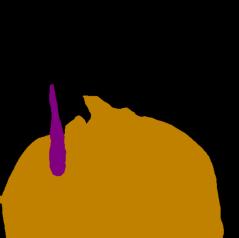}} & 
\adjustbox{valign=c}{\includegraphics[width=0.13\linewidth]{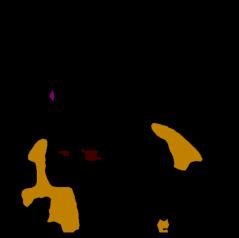}}&  
\adjustbox{valign=c}{\includegraphics[width=0.13\linewidth]{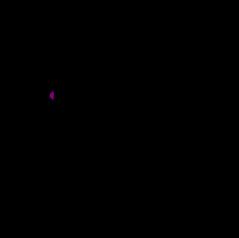}}&
\adjustbox{valign=c}{\includegraphics[width=0.13\linewidth]{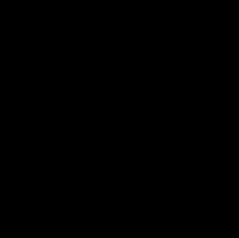}}&
\adjustbox{valign=c}{\includegraphics[width=0.13\linewidth]{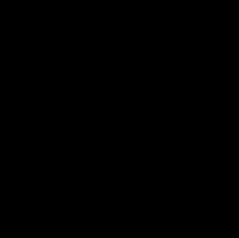}}\\
{ \scriptsize \textbf{IGME} }\hspace{-0.2em} &
\adjustbox{valign=c}{\includegraphics[width=0.13\linewidth]{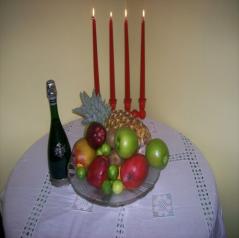}}&  
\adjustbox{valign=c}{\includegraphics[width=0.13\linewidth]{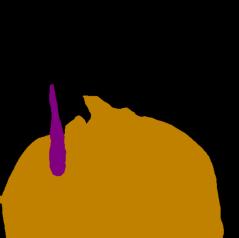}} & 
\adjustbox{valign=c}{\includegraphics[width=0.13\linewidth]{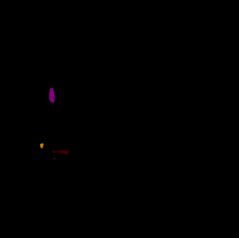}}&  
\adjustbox{valign=c}{\includegraphics[width=0.13\linewidth]{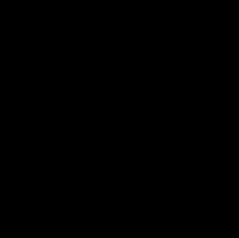}}&
\adjustbox{valign=c}{\includegraphics[width=0.13\linewidth]{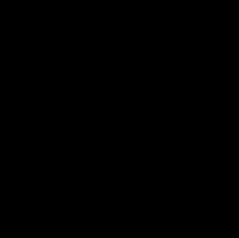}}&
\adjustbox{valign=c}{\includegraphics[width=0.13\linewidth]{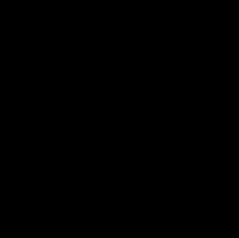}}\\
&\footnotesize{(a)}&\footnotesize{(b)}&\footnotesize{(c)}&\footnotesize{(d)}&\footnotesize{(e)}&\footnotesize{(f)}\\
\end{tabular}
    \caption{\textbf{Visualization of adversarial segmentation results on two examples.} For each example, the baseline and IGME rows show predictions after 1, 5, 10, and 20 attack iterations. (a) is the input image, (b) is the clean segmentation, and (c)--(f) are the attacked predictions after the corresponding iteration counts. These qualitative examples suggest faster prediction degradation for IGME in the shown cases. }
    \label{fig:adversarial_segmentation_results}
\end{figure}

The  Cityscapes comparison in appendix shows a similar mixed pattern: IGME is strongest on the DV3Res101 target in both source blocks, while model ensembles remain competitive on other targets. These results suggest that chained single-source composition can retain adversarial effect across architectures, but its benefit depends on the source target pair.

\begin{table*}[t!]
  \centering
  \footnotesize
  \renewcommand{\arraystretch}{1.0}
  \renewcommand{\tabcolsep}{2.0mm}
  \caption{\textbf{Ablation of attack-component composition strategies.} Bold denotes the best values within each target column.}
  \begin{tabular}{l|c|cc|cc|cc|cc}
  \toprule[1.1pt]
  &&\multicolumn{2}{c|}{Image Quality}&\multicolumn{6}{c}{Target (performance is evaluated with $\text{mIoU}\downarrow$ and $\text{SR}\uparrow$)}\\ \hline
  Source & Attack & $\text{PSNR}\uparrow$ & $\text{SSIM}\uparrow$ & \multicolumn{2}{c|}{DV3Res50} & \multicolumn{2}{c|}{DV3Res101}& \multicolumn{2}{c}{FCNVGG16}  \\ \hline
    \multirow{4}{*}{DV3Res50} 
  & average of perturbation & 32.63 & 0.5492 & 0.0853 & 0.8866 & 0.2624 & 0.6632 & 0.4227 & 0.3507\\ 
   & average of loss & \textbf{33.11} & 0.5356 & 0.0784 & 0.8958 & 0.2568 & 0.6703 & 0.2974 & 0.5432 \\
   & average logit & 32.72 & 0.5466 & \textbf{0.0692} & \textbf{0.9080} & 0.2443 & 0.6864 & 0.2776 & 0.5736 \\
  & chain & 32.82 & \textbf{0.5955} & 0.0710 & 0.9056 & \textbf{0.1985} & \textbf{0.7452} & \textbf{0.2350} & \textbf{0.6390} \\
  \midrule
  \multirow{4}{*}{Mask2Former} 
      & average of perturbation & 32.56 & 0.5650 & 0.5120 & 0.3193 & 0.5430 & 0.3030 & 0.4720 & 0.2750 \\ 
    & average of loss & 32.46 & 0.5793 & 0.4850 & 0.3552 & 0.5120 & 0.3427 & 0.4430 & 0.3195\\
   & average logit & \textbf{32.85} & 0.5822 & 0.4510 & 0.4004 & 0.4780 & 0.3864 & 0.4110 & 0.3687 \\
  & chain & 32.55 & \textbf{0.5949} & \textbf{0.4220} & \textbf{0.4390} & \textbf{0.4610} & \textbf{0.4082} & \textbf{0.3820} & \textbf{0.4132} \\
  \bottomrule[1.1pt]
  \end{tabular}
  \label{tab:method_ablation}
\end{table*}







\begin{table*}[t!]
  \centering
  \footnotesize
  \renewcommand{\arraystretch}{1.0}
  \renewcommand{\tabcolsep}{3.0mm}
  \caption{\textbf{Runtime and transferability comparison on Cityscapes.} 
  The baseline is SVRE, a model-ensemble attack. Time denotes the   average time to generate adversarial examples for one batch under 
  the same image resolution, batch size, and iteration number; the   IGME row uses the default $M=2$ path samples per attack iteration.   Bold numbers indicate the best value within each column, while underlined numbers indicate the second-best value.}
  
  \begin{tabular}{l|ccc|ccc}
  \toprule[1.1pt]
  \textbf{Method} 
  & \multicolumn{3}{c|}{\textbf{DeepLabV3}} 
  & \multicolumn{3}{c}{\textbf{FCN}} \\ \hline
  & \textbf{Time (s)} 
  & \textbf{SR} 
  & \textbf{Trans} 
  & \textbf{Time (s)} 
  & \textbf{SR} 
  & \textbf{Trans} \\ \hline
\textbf{SVRE baseline} 
  & 23.9500 & \textbf{0.6990} & \textbf{0.2023} 
  & 16.9200 & \textbf{0.7810} & \textbf{0.1850} \\ \hline

  \textbf{+ Chained components} 
  & \textbf{3.3496} & 0.6350 & 0.2850 
  & \textbf{2.3593} & 0.6950 & 0.2340 \\ \hline

  \textbf{+ Chained components + IGME} 
  & \underline{7.0500} & \underline{0.6720} & \underline{0.2410} 
  & \underline{4.9700} & \underline{0.7420} & \underline{0.1980} \\

  \bottomrule[1.1pt]
  \end{tabular}
  \label{tab:runtime}
\end{table*}

\noindent\textbf{Runtime trade-off.}
Table~\ref{tab:runtime} separates the effect of chaining from the effect of path averaging. Chained composition is much faster than SVRE because it avoids repeated surrogate-model evaluations. Adding two path samples increases runtime by approximately $2.20\times$ on DeepLabV3 and $2.11\times$ on FCN relative to chaining alone, but recovers part of the SR and transferability gap. The measured ratios are compatible with the $M=2$ design after accounting for batching and fixed tensor overheads.

\noindent\textbf{Update consistency.}
Figure~\ref{fig:cossim_vs} reports cosine similarity between successive update directions on Pascal VOC; the same stabilization trend is observed on Cityscapes. The IGME curve is more consistent in the reported run, supporting the use of path averaging as a step-wise stabilization heuristic. This measurement does not prove that higher cosine similarity always yields stronger transferability. Figure~\ref{fig:adversarial_segmentation_results} provides complementary qualitative examples, where IGME produces faster prediction degradation in the shown cases.

\subsection{Ablation Study}
\noindent\textbf{Effect of path averaging.}
Table~\ref{tab:runtime} shows that the chained baseline provides the lowest runtime, while IGME spends additional computation to improve SR and transferability. This is the intended trade-off: path averaging is retained only when the gain over chaining alone justifies its two-sample cost. A sweep over $M\in\{1,2,3,4\}$ (supplementary material) shows that runtime grows roughly linearly with $M$ as expected, while the target-wise gain from $M>2$ is mixed and does not clearly justify the extra runtime, supporting $M=2$ as the default.

\noindent\textbf{Composition strategy.}
Table~\ref{tab:method_ablation} compares independent averaging strategies with chained composition. Chaining performs strongly in the reported experiment while avoiding repeated branch evaluations. Because the alternatives aggregate at different stages of the attack pipeline, the table should be interpreted as a practical comparison, not as evidence that one operation is an exact approximation of another.


\subsection{On Large Segmentation Model.}
With recent advancement of vision foundation models such as SAM~\cite{kirillov2023segany}, CLIP~\cite{DBLP:conf/icml/RadfordKHRGASAM21}, and ImageBind~\cite{girdhar2023imagebind}, more attention has been paid to transformer-based segmentation models. We perform a preliminary experiment transferring adversarial examples generated on the smallest SAM backbone (ViT-B) to two larger SAM backbones (ViT-L, ViT-H), using the proposed ensemble attack and NI with a noise budget of $12/255$ and 16 iterations. The full setup, qualitative visualizations, and discussion are provided in the supplementary material.

\section{Conclusion}

We presented IGME, a single-source transferable attack framework for semantic segmentation based on chained differentiable attack-component composition and IG-style path-averaged stabilization. Chaining shares the expensive source-model computation across transfer-enhancing components, while the two-sample path average empirically improves step-wise update consistency.

Experiments on Pascal VOC and Cityscapes show that IGME improves several off-diagonal transfer settings relative to strong single-source attacks and offers a favorable runtime trade-off compared with model-ensemble attacks. The gains are not uniform across every architecture, and the SAM result is intentionally reported as a preliminary qualitative study. Overall, the results support efficient attack-component composition as a practical direction for evaluating transferable robustness in dense prediction. Future work will investigate adaptive path-sampling budgets and more systematic transfer evaluation on transformer-based foundation segmentation models.

{
\small
\bibliographystyle{ieee}
\bibliography{egbib}
}

\end{document}